\documentclass[10pt,twocolumn,letterpaper]{article}
\usepackage{graphicx}
\usepackage{subcaption}
\usepackage{graphicx}
\usepackage{amsmath}
\usepackage{amssymb}
\usepackage{amsthm}
\usepackage{subcaption}
\usepackage{color}
\usepackage{authblk}

\usepackage[breaklinks=true,bookmarks=false]{hyperref}

\newcommand{\figref}[1]{\textbf{Fig.\ref{#1}}}
\newcommand{\secref}[1]{\textbf{Sec.\ref{#1}}}
\newcommand{\tableref}[1]{\textbf{Table.\ref{#1}}}
\newcommand{\brackets}[1]{\left(#1\right)}
\newcommand{\loss}{\mathcal{L}}

\newcommand{\E}[1]{\mathrm{E}\left[#1\right]}
\newcommand{\nualpha}{\nu_\alpha}

\newcommand{\ddtheta}[0]{\frac{\partial}{\partial\theta}}
\renewcommand{\thefootnote}{\fnsymbol{footnote}}
\usepackage[symbol]{footmisc}

\title{Closing the gap towards\\end-to-end autonomous vehicle system}

\begin{document}
	\author{Yonatan Glassner\footnote{equal contribution}}
	\author{\hspace{1mm}Liran Gispan$^*$}
	\author{\hspace{1mm}Ariel Ayash$^*$}
	\author{Tal Furman Shohet$^*$}

	\affil{AV AI Solutions - General Motors Israel}

	
	\maketitle
	
\begin{abstract}
Designing a driving policy for autonomous vehicles is a difficult task.
Recent studies suggested an end-to-end (E2E) training of a policy to predict car actuators directly from raw sensory inputs. It is appealing due to the ease of labeled data collection and since hand-crafted features are avoided. Explicit drawbacks such as interpretability, safety enforcement and learning efficiency limit the practical application of the approach. In this paper, we amend the basic E2E architecture to address these shortcomings, while retaining the power of end-to-end learning. A key element in our proposed architecture is formulation of the learning problem as learning of trajectory. We also apply a Gaussian mixture model loss to contend with multi-modal data, and adopt a finance risk measure, conditional value at risk, to emphasize rare events.
We analyze the effect of each concept and present driving performance in a highway scenario in the TORCS simulator.
Video is available in this \href{https://www.youtube.com/watch?v=1JYNBZNOe_4}{link}.
\end{abstract}

	\begin{figure*}[!htb]
		\centering
		\includegraphics[width=0.90\linewidth]{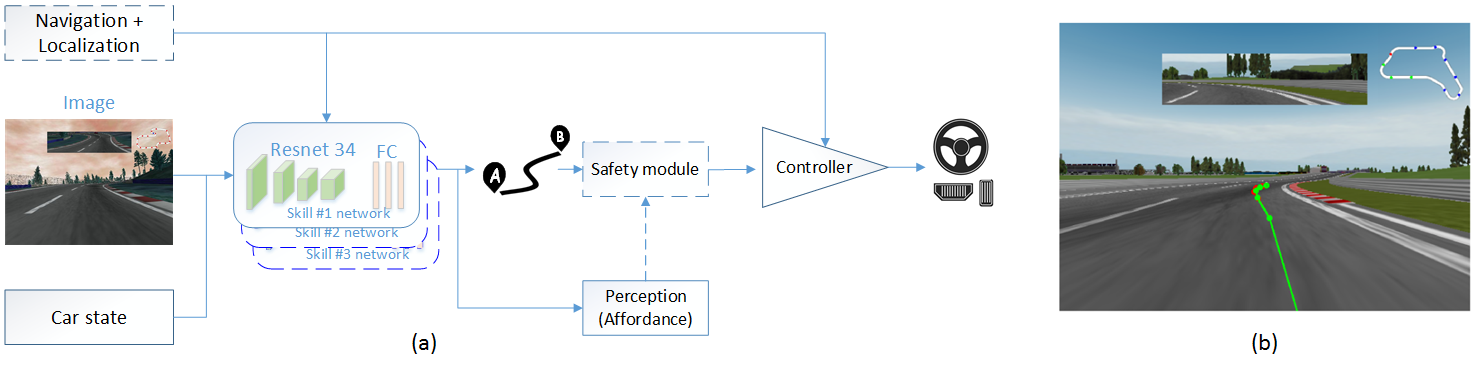}
		\caption{\textbf{Proposed E2E system architecture design and networks output}. (a) The modules of a sub-set architecture employed in this work are enclosed by a solid line. Input image and car state are passed to the network. Trajectory and affordance are predicted simultaneously. Trajectory is provided to the LQR controller, producing low level car actuations. (b) Trajectory predicted by the network projected upon the input image}.\label{fig:figure1}
	\end{figure*}
	\section{Introduction}
"Take me home, car!". The carriage of autonomous transportation has gained significant speed during the last decade, enjoying a powerful tailwind with the blossom of deep learning. However, the rocky-road to obviate the coachmen is still intertwined with challenges to insure the journey remains smooth and safe every time. 
Currently, the two dominating paradigms to autonomous driving are the mediated perception \cite{geiger2013vision,ullman1980against} and the end-to-end (E2E) approaches \cite{pomerleau1989alvinn,bojarski2016end,muller2006off}. 

Mediated-perception paradigm decomposes the task of driving into two salient modules: perception and decision making. The objective of perception is to depict the world state in a meaningful representation. This representation will facilitate the subsequent module to make a decision regarding the appropriate action in the given situation.

Conversely, the end-to-end paradigm suggests to take a more direct approach. In its simplest form, an explicit mapping of the raw inputs into control commands (e.g steering, acceleration, brake)\cite{bojarski2016end} is learned. 
This concept has several advantages: First, supervised training data is easily obtained from recordings of driving demonstrations, avoiding the cumbersome process of data labeling (needed in decomposed approach). Secondly, this approach averts the need for hand-crafted features (i.e., the explicit definition of world state components such as cars, pedestrians, lights, etc.). Those features might omit information useful for decision making. At the same time, they might contain redundant information (e.g., detections which are not necessary for the current decision).                               

Despite the appealing merits, it suffers from several inherent drawbacks:
\\ \textbf{Debuggability and Interpretability}. Direct mapping from input to control presents a challenge for debuggability and interpretability. Evaluating the plausibility of specific actuation command (e.g steering 0.2, acceleration 0.1), given raw data input, is not intuitive. In turn, it hardens the error analysis, thus complicating the development process. Moreover, while driving on a self-driving car, we would like the system to declare its intent. Prediction of car actuators does not provide such declaration.
\\ \textbf{Safety}. Autonomous driving is a life risking task. The enforcement of safety in an E2E system is non-trivial. How can one evaluate the safety of a single actuation command?       
\\ \textbf{Combining localization, map and navigation}. How can the E2E approach effectively leverage information from a map? How can it navigate from a source to destination?   
\\ \textbf{Learning challenges}. Learning driving E2E comprises several big challenges. To name a few:
\\  \textit{Learning efficiency}. It has been argued that compared to the decomposed approach, the E2E paradigm is less efficient in terms of both sample complexity \cite{shalev2016sample} and optimization process \cite{shalev2017failures}.
\\\textit{Rare events}. Suppose you have a highway driving performed by an expert. Majority of the recorded data would be a straight drive. Straightforward E2E learning on this data would likely result in a system incapable of performing well on rare events such as curves, overtakes, etc. This behavior stems from the common deep learning practice of minimizing the average loss. Such optimization procedure might lead the system to be oblivious of high errors caused by rare events. 
 \\\textit{Multi-modal target}. Real life driving decisions in a given scenario are not necessarily consistent. A canonical example is that of left/right overtake of an obstacle \cite{chen2015deepdriving}. This ground-truth multi-modality might be problematic when training a regressor. 

To mitigate these shortcoming, several recent studies proposed to modify the naive E2E approach. Solutions include equipping the E2E system with auxiliary losses related to perception state or high-level (driving) actions \cite{mehta2018learning,xu2017end}, incorporating navigational information as an input to the netwrok \cite{hecker2018end} or training several task-oriented networks \cite{codevilla2018end}.  

Despite this significant progress, it seems that nowadays the E2E approach is still confined to the walls of academic research.

We believe the unfulfilled potential is due to the lack of a system architecture targeting the requirements of real-world driving. This system design should exploit the strengths of the E2E. At same time, it should address the aforementioned shortcomings. In this paper, we depict such system design and present an implementation of several key elements in our architecture, namely:
\\ \textbf{Learning trajectory}. The proposed design learns a sequence of positions (trajectory), leaving low-level actuations to the controller. Prediction of trajectory facilitates better debuggability and interpretability. Trajectory can also be validated by an analytical safety module (e.g., Responsibly-Sensitive sensing \cite{shalev2017formal}), insuring a collision free course. To best of our knowledge, we are the first to formulate the E2E learning problem as one of learning a trajectory. 
\\ \textbf{Focusing on rare events}. Conditional value at risk (CVaR) is a risk measure used in finance and lately adopted for the reinforcement learning setting \cite{tamar2015optimizing}. We adopt this concept to the supervised learning setting, proposing a CVaR loss and metric which focuses on the $\alpha$\% most difficult samples.      
\\ \textbf{Coping with multi-modal target}. We employ a Gaussian Mixture Model (GMM) loss to contend with a multi-modal continuous target \cite{bishop1994mixture}. We characterize several practicalities essential to produce an effective learning of high dimensional target (predicting trajectory) in the context of deep learning.    
\\ \textbf{Leveraging auxiliary loss}. We employ driving affordance \cite{chen2015deepdriving} as an auxiliary loss to our main task of trajectory learning. Affordance serves as a proxy to the full perception state. This auxiliary task contributes to debuggability and interpretability, facilitates safety enforcement and improves learning efficiency.       

We evaluate closed loop driving performance of our end-to-end system in a physical simulator \textbf{TORCS}. Many existing works, implementing E2E approach in the supervised learning setting, report performance on a held-out set only. Such evaluation method does not fully reflect performance in closed-loop driving. We demonstrate smooth high-way driving on test tracks with zero collisions for over 10 hours. 

The main contributions of this paper are:
\\We propose a comprehensive architecture to an E2E system for autonomous driving, addressing the main gaps of the approach.
\\We implement key architecture components, relevant for high-way driving, and present several novelties: 
\begin{itemize}
	\item Learning trajectory instead of car actuators. 
	\item Introduce CVaR loss and metric in the context of deep supervised learning.
	\item Employ GMM loss for high-dimensional target in the context of autonomous driving.
\end{itemize}

	\section{Related work}\label{sec:related_work}
ALVINN is often viewed as the ancestor of nowadays E2E supervised learning approach to autonomous driving \cite{pomerleau1989alvinn}. A simple neural network was applied to map raw input pixels and inputs from a laser range finder directly into desired driving direction. In another fundamental work, \cite{muller2006off} applied a 6 layer convolution network to map input images into steering commands (turn left/right) and avoid obstacless. Despite limited network capacities, they demonstrated the potential of this paradigm.

In a recent seminal work, \cite{bojarski2016end} applied a deep CNN to map images from three front-view cameras to continuous steering commands only, demonstrating control in a limited highway lane following scenario.

Most variants of \cite{bojarski2016end} differ in their inputs (vehicle's speed \cite{yang2018end}, surround view video and route planning information \cite{hecker2018end}, high-level intent \cite{codevilla2018end}) or predicted target (vehichle speed \cite{yang2018end}, high-level intent \cite{xu2017end}, low-level actuation command \cite{codevilla2018end,hecker2018end,bojarski2016end}, or both 
\cite{mehta2018learning}). Others alter their training procedure, specifically by adding different auxiliary losses (driving affordance indicators \cite{mehta2018learning, chen2015deepdriving}, segmentation \cite{xu2017end}, high-level intent \cite{mehta2018learning}).

High-level actions, predicted by \cite{mehta2018learning} and \cite{xu2017end} elucidate the driving intent, 
however, they are too semantic and generating a direct translation to driving is non-trivial. Predicting low-level actions \cite{codevilla2018end, mehta2018learning, yang2018end, hecker2018end} directly translates to driving, 
yet suffers from lack of interpretability and debuggability. 
\cite{mehta2018learning,xu2017end} as well as \cite{hecker2018end} demonstrated their results "in-vitro" on a held-out set only, while actual driving performance remains unknown. 

We relate to the incorporation of high-level intent (e.g., navigation command) as part of our system architecture in \secref{sec:sys_arc}, however its specific implementation is beyond the scope of our current work. Similarly to  \cite{yang2018end}, we provide vehicle speed as an additional input to the network. Conversely, we provide quantitative results of our closed-loop driving performance, measured upon more then $480$ miles of closed-loop high-way driving. We avoid prediction of control commands by inferring future trajectory. This 
output provides an easily explicable depiction of the driving intent, and can be validated for safety and directly applied via controller to perform a closed loop driving   

Focusing training on hard samples, and in particular rare events, has been tackled from several aspects: \cite{lin2018focal} modified the classification cross-entropy loss such that the loss for easy examples is significantly reduced, due to their negligible impact on the accuracy. Their approach cannot be directly applied to a regression problem. \cite{torgo2015resampling} suggests to re-sample the data to directly cope with an imbalanced distribution in the regression setting. However, extending the suggested method to handle high-dimensional targets (as in our case) is not trivial. \cite{shalev2016minimizing} focuses on the difficult examples by minimizing the maximal loss instead of the average loss. Nonetheless, their method might be susceptible to outliers, and does not provide a metric for a regression problem which focuses on the hard examples. \cite{tamar2015optimizing} implemented the CVaR concept in the reinforcement learning setting, by developing a policy gradient algorithm which optimizes the CVaR return instead of the expected return. Using theoretical results of \cite{hong2009simulating}, our work adopted the CVaR concept to the deep supervised learning setting. This approach can be seen as a more holistic approach which modifies \textbf{any} classification or regression loss to focus on the hard examples.

The challenge of coping with multi-modal continuous target has been originally addressed by \cite{bishop1994mixture}, and more recently by \cite{zeldes2017deep}, proposed using a mixture density network (MDN), also known as the GMM loss.  MDN enables to predict an arbitrary conditional probability 
distribution of the target. We successfully apply the GMM loss to a high-dimensional target (applying several necessary practicalities). 
	\section{Building an E2E system}
\subsection{E2E system architecture}\label{sec:sys_arc}
Adequate E2E driving policy has to address: \textit{debugability} and \textit{intepretability, safety, combining localization, map and navigation} and \textit{learning challenges.}
\\In \figref{fig:figure1} we depict the proposed system architecture. Our design is motivated by several guidelines:
\\ \textbf{Learn what is necessary}. Modules such as control, localization and navigation have an established analytical solutions. We believe that a data-driven approach for these modules presents an unnecessary burden on the learning task. Therefore, we exploit such solutions for these modules. The controller utilizes localization signal and translates a trajectory produced by the learning part into low-level actuations. Based on localization on an HD-map, the navigation module triggers an appropriate driving sub-policy - a skill.
\\ \textbf{Skills}. Training a single network to cope with any possible combination of a driving scenarios and a navigation intent is impractical. In reinforcement learning setting, this challenge is tackled by dividing the learned policy into sub-policies - skills \cite{sutton1999between}. In our design, skills are divided according to a combination of navigation command and a (map based) driving scenario (e.g., highway, intersection, etc.). We note that similar to naive E2E, in such a setting it is still easy to obtain supervised data. The only addition is that now we also need to record the localization state and navigator command when data is collected. 
\\ \textbf{Trajectory}. Trajectory prediction is crucial to our design. Opposed to other high-level actions (e.g., cut out/into lane \cite{mehta2018learning}), trajectories can be automatically labeled, and also easily translated into car actuators by a controller. Contrary to direct prediction of car actuators, trajectory declares the network's intent. Hence it is interpretable and debugable. It can also be validated by an analytical safety module. 
\\ \textbf{Perception as auxiliary loss}. Multi-task learning is known to be an effective method to improve learning efficiency \cite{ruder2017overview}. We suggest to learn perception as an auxiliary loss to driving policy. Beyond learning efficiency, this allows better understanding of the network's decisions. We note that adding perception complicates the labeling process. However, we still avoid full dependency between the driving policy (trajectory) and perception state. Hence, data labeling level and scale can be limited. The network can still learn features beyond those defined as perception state.    
\\ \textbf{Safety}. Safety enforcement is essential for autonomous driving. We believe it can be better guarantied via an analytical module (e.g.,\cite{shalev2017formal}), rather than explicitly learned by the system. Our design includes a separate safety module, which can validate the trajectory provided by the network.

In this work, we demonstrate an instantiation of such system design to handle a highway driving
(due to used simulation limitation, we could not simulate intersections nor highway exists). 

In the implemented architecture (see \figref{fig:figure1}), front camera image is provided as an input to the "image backbone network" to extract visual features. These features are then concatenated with the car state and passed through a "sensor fusion network". Resulting representation is used to predict both the trajectory (see \secref{sec:traj}) via the "output network" and the affordance values (auxiliary loss) via the "affordance network" (see \secref{sec:affordance}). We train our system to handle multi-modal targets by using the GMM loss (see \secref{sec:gmm}). In addition, we utilize CVaR loss (\secref{sec:cvar}) to emphasize rare events. On inference, the network predicts a trajectory, which can be validated by a safety module (its implementation was beyond the scope of this work). If the safety module asserts the trajectory is valid - it is followed by the controller. 

We used ResNet as a "backbone network", a stack of fully-connected layers as the "fusion network" and "affordance network". Our "output network" is either linear or GMM layer. We implemented the commonly used LQR controller to follow the trajectory.

\subsection{Learning trajectory}\label{sec:traj}
As described in \secref{sec:sys_arc}, there are several incentives for the network to predict trajectory, rather than raw actuations. We simply model trajectory as a sequence of $K$ points, each with $\brackets{x,y}$ values, in the car-coordinate system. Points along the trajectory are evenly time-spaced. 

As an alternative, we examined the direct prediction of raw-actuations sequence (no controller. network outputs directly applied for driving). This representation provides the option for a safety module. Given the vehicle dynamics, it is possible to analytically translate the sequence of actuation commands into trajectory and validate it. However, we now depend on an accurate knowledge of car-specific vehicle dynamics model (avoided when learning trajectory).
Our experiments show poor results for this approach (It was even worse then single time step prediction \tableref{table:driving_comparison}). This performance gap can be attributed the lack of a feedback loop. When trajectory is predicted, it is followed by a controller. Its internal feedback compensates for the accumulated errors. There is no such mechanism when a sequence of raw actuations is used.  
\subsection{Coping with multi-modal target}\label{sec:gmm}
Consider training data containing situations in which there is an obstacle in front of us. In some of them the taken action will be overtaking it from the right, while in others it would be from the left. Subsequently, for a similiar sensorial input, two different actions are adequate. A simple L2 loss would approximate the optimal solution as moving straight forward, i.e., colliding with the obstacle, while we are interested in a solution which results in an obstacle bypass (left or right) . GMM loss provides such solution. Also known as mixture density network, it was suggested in the seminal work of \cite{bishop1994mixture}. It enables to solve a regression problem, while: 1) coping with a multi-modal target, 2) providing uncertainty , and 3) predicting a conditional probability and not a single value (may be useful for learning a stochastic policy). 

To the best of our knowledge, we are the first to use this approach in the context of a high-dimensional target for a deep learning setting. 

The GMM loss can be written as:
\begin{equation}\label{key}
-\sum_{n=1}^{N}\ln\left[\sum_{k=1}^{K}\pi_{k}\left(x_n,\Theta\right)\cdot\mathcal{N}\left(y_n|\mu_k\left(x_n,\Theta\right),\sigma_{k}^2\left(x_n, \Theta\right)\right)\right]
\end{equation}

where the mixing coefficients $\pi_{k}$  (scalars), the means $\mu_{k}$ (vectors) and the variances $\sigma_{k}^{2}$ (vectors, we assume a diagonal covarriance matrix) are governed by the outputs of the network parametrized by $\Theta$, N is the number of samples in the training data, 
$x_{n}$ is the n'th input sample, $y_{n}$ is the output of sample n, and K is the number of kernels (hyperparameter). 

Intuitively, the model predicts a probability density function (parametrized by $\pi_{k}$, $\mu_{k}$ and $\sigma_{k}^{2}$) for a given input $x_{n}$, and we expect the model to assign a high density to the expert's action $y_{n}$.

The mixing coefficients can be achieved by a softmax operator to satisfy the summation to 1, the variances can be achieved using the exponent operator to satisfy positivity, and the means can be directly taken from the output of a linear layer. Note that despite the diagonal covariance matrix assumption, we do not assume a factorized distribution with respect to the components of the target because of the summation on the kernels.
 
At inference time, the model predicts a distribution of trajectories. We, however, can only follow a single trajectory. We can choose it as the optimal value which maximizes the distribution. This approach will require solving a high-dimensional optimization problem. We take a less computationally demanding approach. We consider the total probability mass function associated with each of the kernels. Trajectory is chosen to be the mean of the largest mixing coefficient, i.e, $\mu_{i_{max}}$ where $i_{max}=\arg\underset{i}{\max}\left(\pi_{i}\right)$. 

GMM loss might lead to large gradients and unstable training, especially for regression problems of high-dimensionality. Hence, we applied several practicalities, among them: 1) Normalization of hidden layers (batch-norm \cite{ioffe2015batch} and predicted targets to diminish the range of losses. 2) Applying log-sum-exp trick for numerical stability. 3) Gradient clipping.  4) Running "warm-up" epochs while freezing $\sigma_i$ before "opening" it for learning.

\subsection{Focusing on rare events}\label{sec:cvar}
In every supervised learning (SL) problem there are easy and hard samples. One way of quantifying a sample's hardness is by its loss. This served as an incentive for various loss-oriented sampling policies (\cite{shalev2016minimizing}) ans loss modifications (\cite{lin2018focal}) .

A well known risk measure, extensively researched in the finance domain (i.e. \cite{rockafellar2000optimization}), is the conditional value-at-risk (CVaR). For a given random variable $\loss$, the $\alpha$-CVaR is defined as the conditional expectation of $\loss$, over the $\alpha\%$ highest values:
\begin{equation}
CVAR_\alpha(\loss)=E\left[\loss|\loss>\nu_\alpha\right] 
\label{def:cvar}
\end{equation}
where $\nualpha$ is the $\alpha$-upper-quantile of $\loss$.

Within the SL framework, we can consider $\loss$ to be the loss of a given model, defined by a set of parameters $\theta$, 	over a given sample distribution $\mathcal{D}$. 

In the context of autonomous driving, we are interested in diminishing (scarce) large errors, possibly corresponding to a fatal behavior. At the same time, we would like to avoid over-fitting (prevalent) small errors. Thus, instead of minimizing the expected loss, we would like to minimize the CVaR. We are interested in a term for $\nabla_\theta CVaR_{\alpha}\left(\loss\right)$. 
\\Using Theorem 3.1 of \cite{hong2009simulating}, the gradient of the CVaR 
is given by:

\begin{equation}
\ddtheta CVaR_\alpha\left[\loss\right]=\E{\ddtheta \loss \Bigg\vert \loss>\nualpha}
\label{cvar_grad}
\end{equation}

Using \eqref{cvar_grad}, we can apply stochastic gradient descent to optimize the CVaR loss. The CVaR gradient may be estimated by sampling a batch, passing it through the network, obtaining the per-sample loss, calculating the empirical  $\hat{\nu}_\alpha$, and back-propagating the loss gradient only for samples with a loss value above $\hat{\nu}_\alpha$.

To the best of our knowledge, we are the first to apply CVaR in the context of deep stochastic optimization. We next show applying this loss significantly improves the system performance on rare events.
In our experiments (see \secref{sec:results}), we demonstrate that this gradient estimation optimizes the CVaR for both train and validation set. The outcome is a risk-averse agent, trading-off its performance on common driving scenarios with a more safe behavior in rare events.

Note, that besides functioning as an optimization algorithm, we use the CVaR as an evaluation metric to assess the worst case behavior. We encourage the use of CVaR either as a loss or as a metric for SL regression problems.

\subsection{Leveraging auxiliary loss}\label{sec:affordance}
Our proposed architecture design employs the learning of perception state as an auxiliary loss. As a proxy to the perception state we utilize driving affordance. Our set of affordance indicators is similar to the one originally proposed by \cite{chen2015deepdriving}, consisting of self heading angle relative to road tangent and a set of distances from adjacent cars and lane markings. The multi-task loss function is a weighted sum of trajectory and affordance losses, where the weights value found through grid-search.         
	\section{Experiments}\label{sec:results}
\subsection{Experimental settings}\label{sec:exp_set}
To demonstrate our approach, we utilize TORCS 3D car racing simulator \cite{wymann2000torcs}.
We extract simulated front-camera images provided by the game (rear-view mirror is cropped out). Frames are saved to RAM by redirecting the screen rendering into OpenGL texture, sending it through TCP-IP socket. We also save the relevant state of the vehicle (position and velocity), as well as distances from other vehicles in the scene. 
\subsubsection{The expert}

To perform behavioral cloning, one needs an expert. The simplest option is to
record a human player whilst driving a remote agent. Despite being a decent target 
for imitation, collecting a large body of annotated data in such fashion becomes an exhausting process.

One alternative is programming a rule-based agent. 
The existing TORCS agents were found unsuitable, as they neither obey traffic rules nor implement safety considerations.
Instead, we have designed a basic expert traversing the inner states of a finite state machine (FSM). 

The expert adheres to the following guide lines: 1. The velocity will not exceed an upper limit, defined according to road curvature and friction. 2. A basic controller is employed to keep the vehicle in the center of the lane. 3. States of the FSM define when the expert \textit{can} and when he \textit{should} overtake. 

The expert's behavior is randomized in terms of overtake initiation and finish distances to simulate multi-modal behavior exhibited by real drivers. Adding noise to the expert was found to be crucial for a proper state space exploration.

We collected ~500K samples, used for network training (70\%) and validation (30\%).

\subsubsection{The agent}

Apart from quantitative performance on a held-out set, we evaluate the driving skills of our agent in the \textbf{TORCS} simulation, over unseen tracks scenarios.

An image from a front facing camera, along with the vehicle state is captured at 50 fps. This data is sent via socket to our pre-trained agent. A 1.5 sec ahead trajectory is computed every 100ms, by performing a forward pass of the image and current velocity through the network. Output trajectory is provided to 
a low-level controller which follows it by sending actuators commands back to the simulator. Similarly to real world conditions, there is no synchronization mechanism between the simulator and our agent.

We test our agent both on an empty track and in scenes with static obstacles. In the latter, we randomly place 20 to 80 parking vehicles on the track, in intervals of 50 to 200 meters. Recorded tracks were devided to train and validation (the 70\% of samples used for training contained no images from validation tracks).

\subsection{Results}\label{sec:results_traj}
Our network learns to predict the vehicle's trajectory over a period of 1.5 seconds into the future. This trajecorty is sampled at 300 mSec intervals, providing a total of 5 data points ($(x,y)$ pairs). 

Our input is a VGA image (480x640). The sky were cropped out, leaving a 320x640 input image. We use a Resnet-34 for the "image backbone network", a 3-layers fully connected network (each with 512 hidden neurons) for the "sensor fusion network", and additional $512\times10$ linear layer to predict the output $\left( \textit{5 points} \times \textit{2 dimension each}\right)$. We use Adam optimizer \cite{kingma2014adam} (learning rate $10^{-3}$, no weight decay). All fully-connected layers are trained with dropout 0.5 \cite{ioffe2015batch}. We found it useful to normalize the predicted values (zero-mean and unit variance) as a pre-process. Otherwise, the prediction of (large value) longitudinal positions is vastly dominant to that of the (small value) lateral. So is the case for prediction of the (large) future positions, superior to that of the (small) near-future positions.

We train our network with a GMM loss (see \secref{sec:gmm}). The GMM layer consists of two modes. Each mode has 10 parameters for the mean $\left(\vec{\mu}_i\right)$, 10 parameters for variance $\left(\vec{\sigma}_i\right)$ and an additional per-mode parameter $\left(\pi_i\right)$, resulting in network output size of 42. Other network parameters remain similar. To avoid convergence and numerical instability issues (due to the high-dimensional target), we first freeze the $\sigma_i$ for 5 epochs and then "open" all GMM parameters for training. 

\begin{figure}[h!]
	\centering
	\includegraphics[trim={1cm 1cm 0 0  },clip,width=\linewidth]{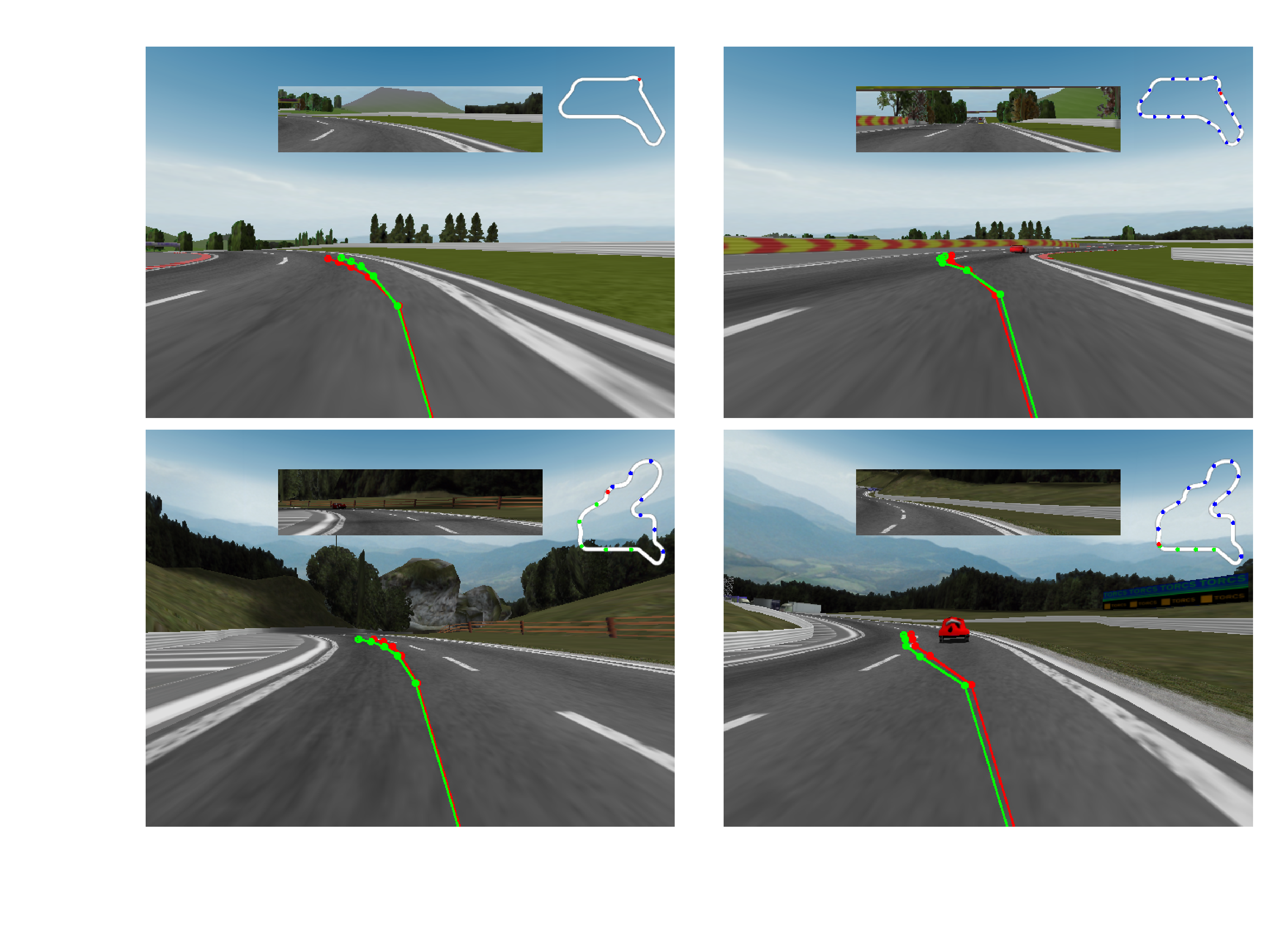}
	\caption{\textbf{Output trajectory examples for multiple scenarios.} The ground truth is marked with green line. The predicted trajectory is marked with red line.}
	\label{fig:output_traj_example}
\end{figure}

\begin{figure}[h!]
	\centering
	\begin{subfigure}[b]{0.45\linewidth}
		\includegraphics[trim={0 2cm 0 0}, clip,width=\linewidth]{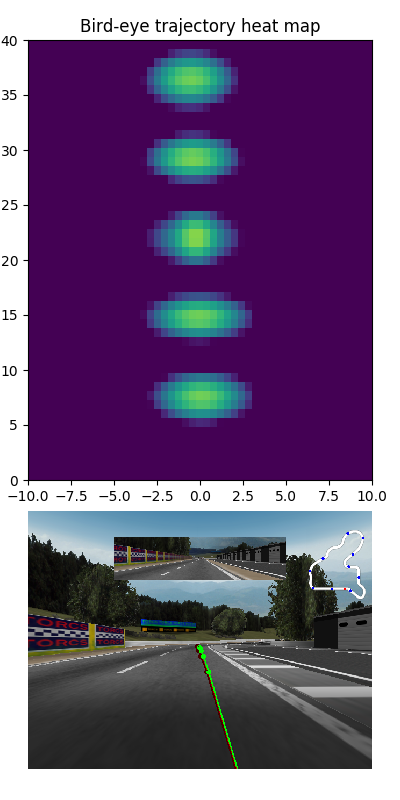}
		\caption{Unimodal prediction}
						\label{fig:high_conf}
	\end{subfigure}
	\begin{subfigure}[b]{0.45\linewidth}
		\includegraphics[trim={0 2cm 0 0}, clip, width=\linewidth]{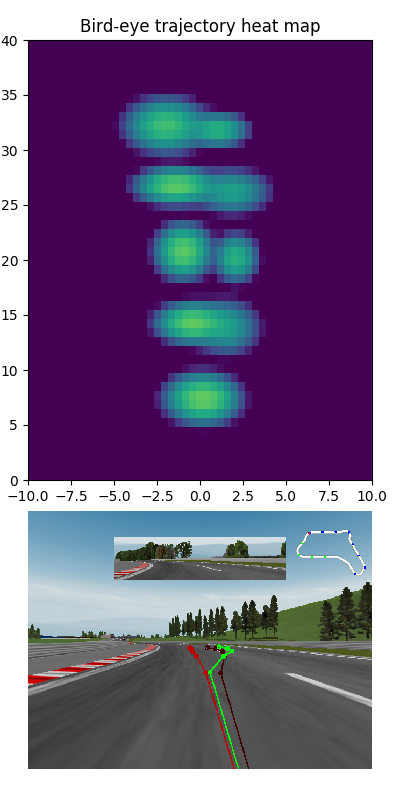}
				\caption{Bi-modal prediction}
				\label{fig:low_conf}
	\end{subfigure}
	\caption{\textbf{Trajectory prediction provided by GMM layer}. The left figure demonstrates an "easy" unimodal decision sample, where going straight is the only rational option. In contrast, the right figure demonstrates a multi-modal prediction, where the system hesitates between completing an overtake and returning right, to staying on the left lane. }
	\label{fig:confidence}
\end{figure}

Typical qualitative results, in various road segments and expert's poses, are given in \figref{fig:output_traj_example} and \figref{fig:confidence}. In \figref{fig:output_traj_example} the ground-truth expert's trajectory is compared to the predicted one (the maximal mode). These results show clearly that the trajectory formulation can be well learned.
The multi-modal effect is exemplified in \figref{fig:confidence}. In the left figure (\figref{fig:high_conf}) one can see a "keep straight" example. Both GMM modes collapse into a uni-modal prediction. Conversely, the right image (\figref{fig:low_conf}), depicts an example for which multiple options are available: finalize overtake now, or continue and finalize the overtake later. Accordingly, the GMM outputs now predict two legitimate actions.

We evaluate our agent over unobserved \textit{validation routes}, varying number of opponents and their positions. Our agent drives smoothly in an average speed of 48 mile per hour, with 12.7 collisions per 100 mile on average (see \tableref{table:driving_comparison}).

We employ affordance as an auxiliary loss (see \secref{sec:affordance}). We train the network to predict both trajectory and affordance values. 
Beyond its contribution to the interpretability of the system, its addition leads to collision rate improvement (\tableref{table:driving_comparison}) - 0.83 collisions per 100 mile instead of 12.7 (see \tableref{table:driving_comparison}).

As aforementioned, training deep network for the autonomous vehicles' domain, requires coping with imbalanced data. Our dataset mostly consisting of straight segments, with limited number of overtakes (similarly to a real-world high-way driving distribution).

Unsurprisingly, we observe that using a random sampler leads the network to rapidly adapt to these scenarios, presenting poor results on the less prevalent cases (e.g., curved segments or static cars on track). This behavior is reflected in a high-loss value for these rare cases, numerically-observed by a high CVaR metric (see \secref{sec:cvar}). 

\begin{figure}[h!]
	\centering
	\begin{subfigure}[b]{\linewidth}
			\includegraphics[trim={0.5cm 0.6cm 0 0  },clip,width=\linewidth]{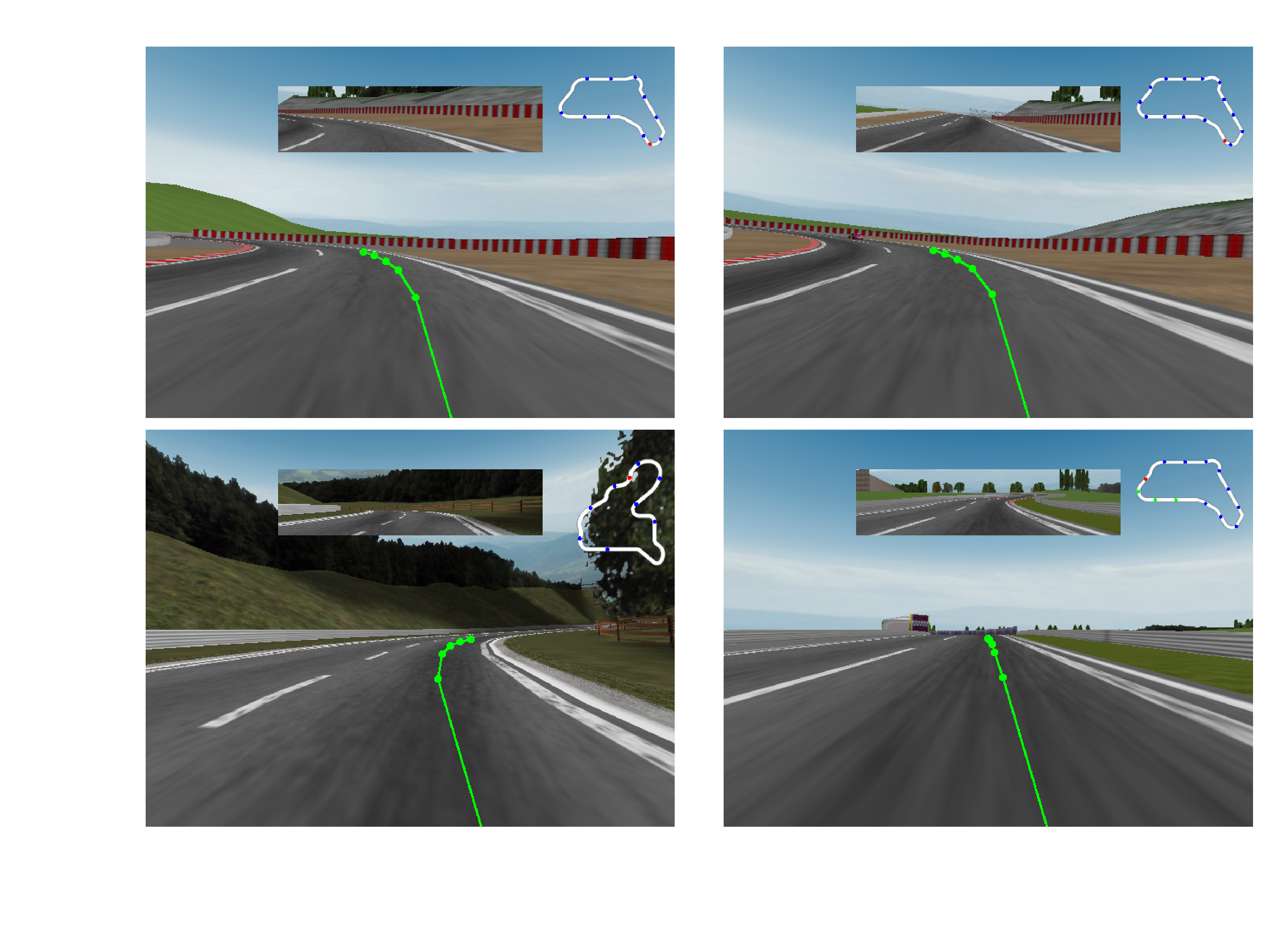}
		\caption{\textbf{Random sampling examples.} Samples are simple straight trajectories for their high prevalence.}
		\label{fig:random_sampling}
	\end{subfigure}

	\begin{subfigure}[b]{\linewidth}
	\includegraphics[trim={0.5cm 0.6cm 0 0  },clip,width=\linewidth]{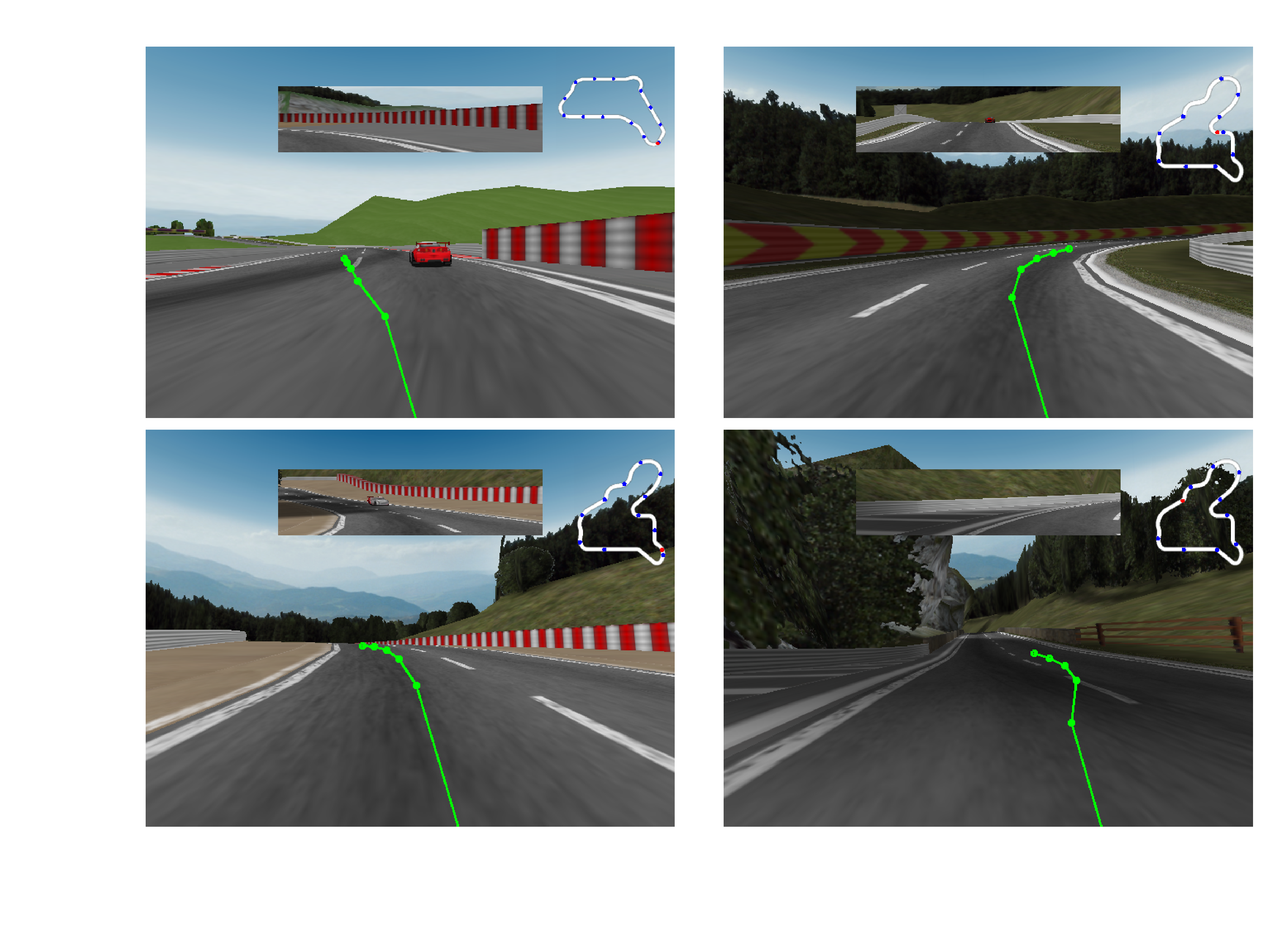}
	
	\caption{\textbf{CVAR loss examples.} Samples contains a variety of situations, including curves, overtakes, etc.}
		\label{fig:cvar_sampling}
	\end{subfigure}
	\caption{\textbf{CVaR vs. random sampling effect.}}
	\label{fig:cvar_vs_random}
\end{figure}

To cope with this phenomenon, we fine-tune the network with \textit{CVaR loss}, emphasizing these hard cases. In our experiments, one additional epoch of training was sufficient to demonstrate the difference between an average-loss and a CVaR loss. Note, that with random sampler, most images present a very similar scene: straight road. CVaR loss, on the other hand, optimizes among various scenes: straight and curved segments, with and without cars (see \figref{fig:cvar_vs_random}).

\begin{figure}[h!]
	\centering
	\begin{subfigure}[b]{\linewidth}
		\includegraphics[width=\linewidth]{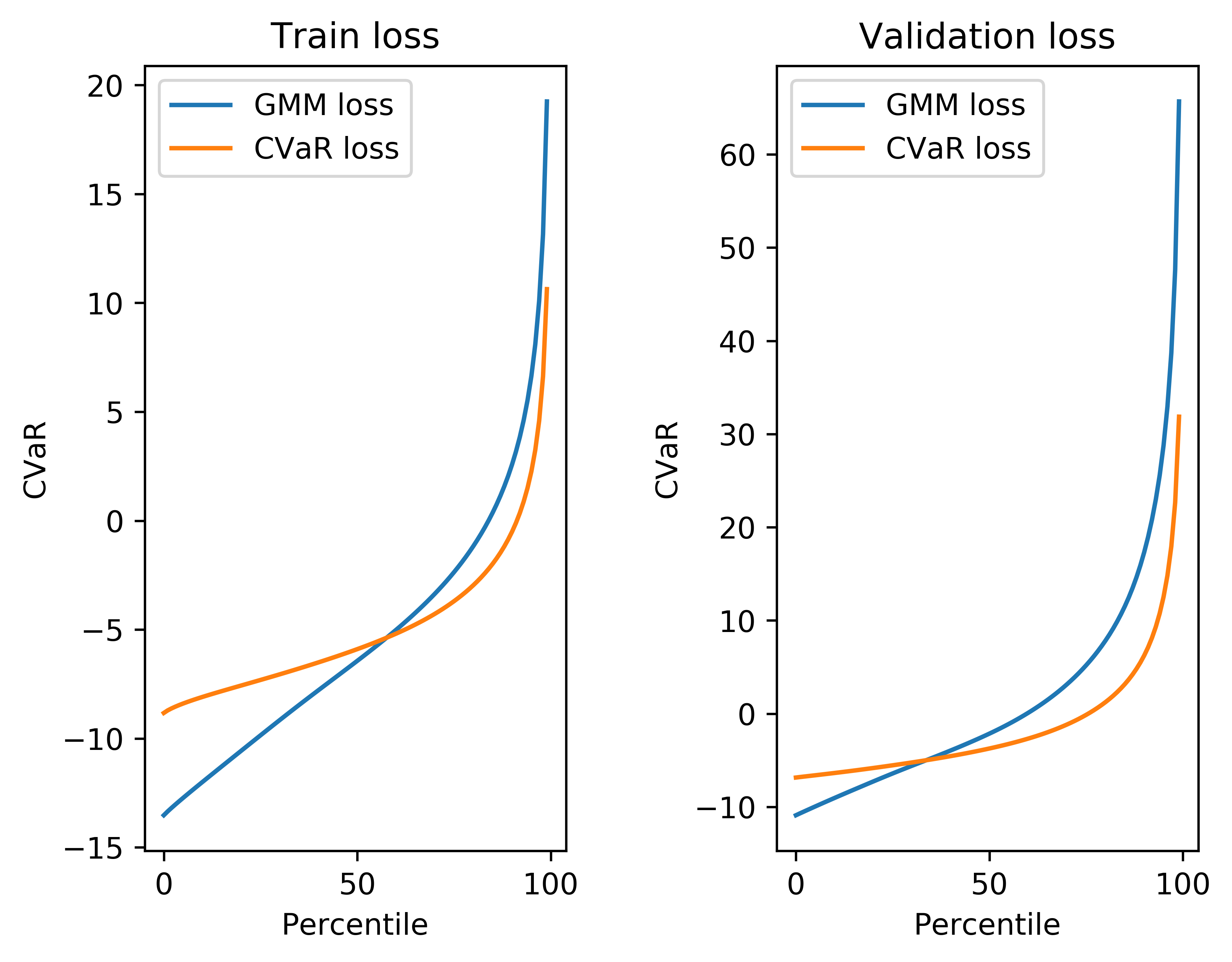}
	\end{subfigure}
	\caption{CVaR-per-percentile for regular GMM-loss and CVaR loss. CVaR effect is exemplified- with decreased CVaR for the hard cases.}
	\label{fig:cvar_results}
\end{figure}

To demonstrate the CVaR-metric, we use it to compare results of avarage-loss and CVaR loss network training. We measure the CVaR per each percentile (lower is better). For CVaR loss optimization, we use CVaR-90, i.e., optimizing the loss values above 90th percentile (see \figref{fig:cvar_results}).

Despite a moderate decrease in performance on the easier cases (the orange line is above the blue line on the left side), the CVaR loss achieves lower values on the hard cases (the orange line is lower than the blue line in the right side), leading to an overall better and safer driving performance. This effect has generalized for the validation set as well. 

This result also reflects in the driving performance. When trained without CVaR-loss, our agent achieved $0.83$ collisions per 100 mile, while with a CVaR loss, the agent covered the same distance with \textbf{zero collisions} (see \tableref{table:driving_comparison}).

All collision rates are summarized in \tableref{table:driving_comparison}. As a baseline we implemented an E2E agent, following the concept of \cite{bojarski2016end}. Contrast to the original work, we used Resnet-34 instead of Pilotnet and predicted all actuation commands instead of steering-only. This approach achieved poor results, with 183.3 collision per 100 miles. Moving to trajectory diminished the collision rate to 12.7. The addition of affordance and CVaR diminishes the collision rate to \textit{zero}.

\begin{table}[h!]
	\centering
	\begin{tabular}{|c|c|c|}
		\hline 
		Method & \shortstack{$\#$ Collisions per 100 mile} \\ \hline 
		\small{Baseline E2E (\cite{bojarski2016end})}  & 183.3  \\ \hline 
		\small{Trajectory GMM} & 12.7 \\ \hline 
		\small{GMM + affordance}  & 0.83 \\ \hline 
		\small{GMM + affordance + CVaR} & \textbf{0}\\ \hline 
	\end{tabular} 
	\caption{\textbf{Ablation study - collision rate}. Our final agent achieved zero collisions rate.}
	\label{table:driving_comparison}
\end{table}

	\section{Summary}
In this work we introduce an E2E system architecture for autonomous driving. This architecture aims to exploit the feature learning and data collection benefits of E2E learning, while keeping the system interpretable and safe. A key enabler is the formulation of the learning problem as learning of trajectory. 

We implement key architecture components, relevant for high-way driving. To contend with the challenges arising from closed-loop driving, we apply a Gaussian mixture model loss to cope with multi-modal high-dimensionality targets and the conditional value at risk concept to emphasize rare events. 

We analyze the contribution of each modification, and demonstrate smooth high-way driving with zero collisions in TORCS simulator. 

While the presented design and its application to E2E closed-loop driving are encouraging, it is clear that further effort is needed to enable comprehensive real-world driving.  

Generalization of the approach towards full dynamic scene using multiple sensors and temporal inputs, incorporating localization, navigation and safety modules, and learning multiple skills are important directions for future work.

	\clearpage
	
	{\small
		\bibliographystyle{ieee}
		\bibliography{ref}

\begin{thebibliography}{10}\itemsep=-1pt

\bibitem{bishop1994mixture}
C.~M. Bishop.
\newblock Mixture density networks.
\newblock Technical report, Citeseer, 1994.

\bibitem{bojarski2016end}
M.~Bojarski, D.~Del~Testa, D.~Dworakowski, B.~Firner, B.~Flepp, P.~Goyal, L.~D.
  Jackel, M.~Monfort, U.~Muller, J.~Zhang, et~al.
\newblock End to end learning for self-driving cars.
\newblock {\em arXiv preprint arXiv:1604.07316}, 2016.

\bibitem{chen2015deepdriving}
C.~Chen, A.~Seff, A.~Kornhauser, and J.~Xiao.
\newblock Deepdriving: Learning affordance for direct perception in autonomous
  driving.
\newblock In {\em Proceedings of the IEEE International Conference on Computer
  Vision}, pages 2722--2730, 2015.

\bibitem{codevilla2018end}
F.~Codevilla, M.~Miiller, A.~L{\'o}pez, V.~Koltun, and A.~Dosovitskiy.
\newblock End-to-end driving via conditional imitation learning.
\newblock In {\em 2018 IEEE International Conference on Robotics and Automation
  (ICRA)}, pages 1--9. IEEE, 2018.

\bibitem{geiger2013vision}
A.~Geiger, P.~Lenz, C.~Stiller, and R.~Urtasun.
\newblock Vision meets robotics: The kitti dataset.
\newblock {\em The International Journal of Robotics Research},
  32(11):1231--1237, 2013.

\bibitem{hecker2018end}
S.~Hecker, D.~Dai, and L.~Van~Gool.
\newblock End-to-end learning of driving models with surround-view cameras and
  route planners.
\newblock In {\em European Conference on Computer Vision (ECCV)}, 2018.

\bibitem{hong2009simulating}
L.~J. Hong and G.~Liu.
\newblock Simulating sensitivities of conditional value at risk.
\newblock {\em Management Science}, 55(2):281--293, 2009.

\bibitem{ioffe2015batch}
S.~Ioffe and C.~Szegedy.
\newblock Batch normalization: Accelerating deep network training by reducing
  internal covariate shift.
\newblock {\em arXiv preprint arXiv:1502.03167}, 2015.

\bibitem{kingma2014adam}
D.~P. Kingma and J.~Ba.
\newblock Adam: A method for stochastic optimization.
\newblock {\em arXiv preprint arXiv:1412.6980}, 2014.

\bibitem{lin2018focal}
T.-Y. Lin, P.~Goyal, R.~Girshick, K.~He, and P.~Doll{\'a}r.
\newblock Focal loss for dense object detection.
\newblock {\em IEEE transactions on pattern analysis and machine intelligence},
  2018.

\bibitem{mehta2018learning}
A.~Mehta, A.~Subramanian, and A.~Subramanian.
\newblock Learning end-to-end autonomous driving using guided auxiliary
  supervision.
\newblock {\em arXiv preprint arXiv:1808.10393}, 2018.

\bibitem{muller2006off}
U.~Muller, J.~Ben, E.~Cosatto, B.~Flepp, and Y.~L. Cun.
\newblock Off-road obstacle avoidance through end-to-end learning.
\newblock In {\em Advances in neural information processing systems}, pages
  739--746, 2006.

\bibitem{pomerleau1989alvinn}
D.~A. Pomerleau.
\newblock Alvinn: An autonomous land vehicle in a neural network.
\newblock In {\em Advances in neural information processing systems}, pages
  305--313, 1989.

\bibitem{rockafellar2000optimization}
R.~T. Rockafellar, S.~Uryasev, et~al.
\newblock Optimization of conditional value-at-risk.
\newblock {\em Journal of risk}, 2:21--42, 2000.

\bibitem{ruder2017overview}
S.~Ruder.
\newblock An overview of multi-task learning in deep neural networks.
\newblock {\em arXiv preprint arXiv:1706.05098}, 2017.

\bibitem{shalev2017failures}
S.~Shalev-Shwartz, O.~Shamir, and S.~Shammah.
\newblock Failures of gradient-based deep learning.
\newblock {\em arXiv preprint arXiv:1703.07950}, 2017.

\bibitem{shalev2017formal}
S.~Shalev-Shwartz, S.~Shammah, and A.~Shashua.
\newblock On a formal model of safe and scalable self-driving cars.
\newblock {\em arXiv preprint arXiv:1708.06374}, 2017.

\bibitem{shalev2016sample}
S.~Shalev-Shwartz and A.~Shashua.
\newblock On the sample complexity of end-to-end training vs. semantic
  abstraction training.
\newblock {\em arXiv preprint arXiv:1604.06915}, 2016.

\bibitem{shalev2016minimizing}
S.~Shalev-Shwartz and Y.~Wexler.
\newblock Minimizing the maximal loss: How and why.
\newblock In {\em ICML}, pages 793--801, 2016.

\bibitem{sutton1999between}
R.~S. Sutton, D.~Precup, and S.~Singh.
\newblock Between mdps and semi-mdps: A framework for temporal abstraction in
  reinforcement learning.
\newblock {\em Artificial intelligence}, 112(1-2):181--211, 1999.

\bibitem{tamar2015optimizing}
A.~Tamar, Y.~Glassner, and S.~Mannor.
\newblock Optimizing the cvar via sampling.
\newblock In {\em AAAI}, pages 2993--2999, 2015.

\bibitem{torgo2015resampling}
L.~Torgo, P.~Branco, R.~P. Ribeiro, and B.~Pfahringer.
\newblock Resampling strategies for regression.
\newblock {\em Expert Systems}, 32(3):465--476, 2015.

\bibitem{ullman1980against}
S.~Ullman.
\newblock Against direct perception.
\newblock {\em Behavioral and Brain Sciences}, 3(3):373--381, 1980.

\bibitem{wymann2000torcs}
B.~Wymann, E.~Espi{\'e}, C.~Guionneau, C.~Dimitrakakis, R.~Coulom, and
  A.~Sumner.
\newblock Torcs, the open racing car simulator.
\newblock {\em Software available at http://torcs. sourceforge. net}, 4:6,
  2000.

\bibitem{xu2017end}
H.~Xu, Y.~Gao, F.~Yu, and T.~Darrell.
\newblock End-to-end learning of driving models from large-scale video
  datasets.
\newblock {\em arXiv preprint}, 2017.

\bibitem{yang2018end}
Z.~Yang, Y.~Zhang, J.~Yu, J.~Cai, and J.~Luo.
\newblock End-to-end multi-modal multi-task vehicle control for self-driving
  cars with visual perception.
\newblock {\em arXiv preprint arXiv:1801.06734}, 2018.

\bibitem{zeldes2017deep}
Y.~Zeldes, S.~Theodorakis, E.~Solodnik, A.~Rotman, G.~Chamiel, and D.~Friedman.
\newblock Deep density networks and uncertainty in recommender systems.
\newblock {\em arXiv preprint arXiv:1711.02487}, 2017.

\end{thebibliography}
	}
	
\end{document}